\pdfoutput=1
\documentclass[10pt, a4paper]{article}
\usepackage[final]{lrec2026}
\usepackage{booktabs}
\usepackage{amsmath}
\usepackage{enumitem}

\newcommand{\tool}[1]{\texttt{#1}}

\title{A Corpus-Aligned Uthmani-to-Standard Quranic Word Mapping and a Deterministic Recitation Validator}

\name{Yahya Mohamed Elnawasany}
\address{Independent Researcher, Egypt \\
         \texttt{yahyaalnwsany39@gmail.com}}

\abstract{
Quranic text is distributed in two orthographic forms that are byte-level distinct: the Uthmani script used in every printed mushaf, and the Standard (Imla'i) Arabic form that every mainstream Arabic NLP tool is built for. The gap is concentrated in one Unicode character, U+0670 (superscript alef), which appears in some of the most frequently recited words in the Quran and is silently mishandled by general-purpose Arabic normalizers. We release a 2{,}290-pair, corpus-aligned Uthmani-to-Standard word mapping constructed by aligning the complete 6{,}236-verse Quran across both orthographic forms, together with a seven-step text normalization pipeline built on it. Normalizing both forms of all 6{,}236 verses through that pipeline yields identical strings for 90.9\% of verses, and we characterize the residual divergence rather than assert that it is closed. On top of the normalized text, we build a deterministic, LLM-free Quranic recitation validator using a four-layer verse-matching search (exact, morphological, relaxed, fuzzy) and word-error-rate-graded feedback across five severity tiers. The validator scores 98.4\% (122/124) on a 124-case suite emitted by the released test harness, and both failures share one mechanism: a single substitution error can make a different verse an exact match. A full-corpus census additionally quantifies an inherent text-only ambiguity affecting 16.5\% of verses, and on 34 recitation transcripts drawn from a deployed Arabic ASR system the validator identifies the correct verse in every case. We release the mapping, the script that builds it, the validator, and the evaluation harness under open licenses; every number in this paper except the deployment measurement, whose transcripts are not ours to publish, is reproduced by running them.
}

\begin{document}
\maketitleabstract

\section{Introduction}
\label{sec:intro}

Quranic recitation validation -- determining whether spoken text matches the canonical Quran and, if not, where -- requires comparing automatic-speech-recognition (ASR) output against an indexed reference corpus. This is harder than it appears for a specific, underreported reason: the Quran exists in two orthographic forms that are byte-level distinct even when phonetically identical. The \emph{Uthmani} script, standardized in the era of the third Caliph and used in every printed mushaf and in authoritative digital corpora such as Tanzil \citeplanguageresource{tanzil2024}, includes characters absent from \emph{Standard} (\emph{Imla'i}) Arabic -- the form every mainstream Arabic NLP tool, ASR model, and text corpus is built around. The single most consequential of these characters is U+0670, the Arabic Superscript Alef (\emph{alef khanjariyya}), which Uthmani writes as a diacritic above the preceding letter. What Standard writes in its place is not fixed, and that is the difficulty: for most of these words it writes a full alef, so that Uthmani \emph{al-kafirin} gains one; but for a sizeable minority it writes nothing at all, and \emph{al-rahman} -- among the most frequently recited words in the Quran -- is spelled without an alef in both forms. A single global rule is therefore wrong in one direction or the other whichever way it is written, which is what motivates a word-level resource rather than another character-level heuristic. To our knowledge, no published word-level resource maps between the two forms, and the normalization routines shipped with mainstream Arabic NLP toolkits -- Farasa \cite{abdelali-etal-2016-farasa}, CAMeL Tools \cite{obeid-etal-2020-camel} and PyArabic \cite{zerrouki2023pyarabic} -- treat U+0670 as an ordinary diacritic: they either strip it silently, losing the long vowel it represents, or leave it untreated, in both cases producing byte-level mismatches between Uthmani-sourced text and Standard-form indexes or ASR output.\footnote{Measured on the 2{,}290 released word pairs (\S\ref{sec:dataset}) under Python 3.12 with PyArabic 0.6.15 and CAMeL Tools 1.6.0. \tool{pyarabic.araby.strip\_tashkeel} leaves U+0670 in place in all 2{,}290 words; \tool{pyarabic.araby.strip\_diacritics} and CAMeL Tools' \tool{camel\_tools.utils.dediac.dediac\_ar} delete it in all 2{,}290, which yields the correct Standard form for only 359 (15.7\%). The two behaviours are not interchangeable errors: Standard orthography writes a full alef where Uthmani has U+0670 in 1{,}335 of the 2{,}290 words and writes nothing in 362, the remainder differing for additional orthographic reasons, so neither a global delete nor a global keep is correct. Farasa is cited here as a mainstream toolkit; we report measurements only for the two Python toolkits we ran.}

This paper makes two contributions. First, a 2{,}290-pair, corpus-aligned Uthmani-to-Standard word mapping (\S\ref{sec:dataset}), constructed by aligning the complete Quran across both orthographic forms -- to our knowledge the first published resource of its kind. Second, a deterministic recitation validator built on top of it (\S\ref{sec:validator}), which normalizes ASR output through a seven-step pipeline anchored by the mapping and identifies the recited verse via a four-layer search, scoring 98.4\% on a 124-case evaluation suite (\S\ref{sec:eval}) with no language model involved anywhere in the pipeline. Both the dataset and the validator's code are released publicly (\S\ref{sec:availability}).

\section{Related Work}
\label{sec:related}

Digital Quranic text resources are well established: the Tanzil project \citeplanguageresource{tanzil2024} maintains the canonical Uthmani and Simple-Arabic encodings used as ground truth throughout this work, and the Quranic Arabic Corpus \citeplanguageresource{quranicarabiccorpus} provides morphological annotation (root, lemma, part of speech) for every word, which our validator's linguistic-matching layer (\S\ref{sec:validator}) draws on. Neither resource, however, provides an explicit word-level alignment between the Uthmani and Standard orthographic forms -- the specific gap this paper addresses. The problem of mapping a variant written form onto a single conventional one is itself well established in Arabic NLP: CODA \cite{habash-etal-2012-coda} defines a conventional orthography for dialectal Arabic precisely so that dialectal text becomes tractable for tools built for Modern Standard Arabic. Our task has the same shape but a different character. CODA is \emph{prescriptive} -- it stipulates a convention for varieties that have none, and is defined over undiacritized text -- whereas the Uthmani and Standard forms are both already standardized and fully diacritized, so the correspondence between them is \emph{descriptive} and recoverable by alignment over parallel corpora rather than stipulated by guideline. General Arabic NLP surveys \cite{habash2010arabic,darwish2014arabic} document root-based morphology and diacritization as central Arabic NLP challenges, and diacritization-restoration systems \cite{zerrouki2019tashkeel} address the general \emph{absence} of diacritics in Modern Standard Arabic text; the Uthmani/Standard divide is a distinct problem, since Quranic text is already fully diacritized in both forms and the mismatch is orthographic rather than a missing-information problem. Prior work on computational recitation assessment \cite{alqahtani2019quran_nlp} surveys two dominant approaches -- forced phoneme alignment and end-to-end acoustic classification -- both of which require a Tajweed-aware acoustic model or a large labeled-error audio corpus, neither of which exists at production quality for Uthmani-script Quranic recitation. The closest adjacent effort is Tarteel \cite{khan2021tarteel}, which releases a large crowd-sourced corpus of Quranic recitation audio paired with its text and underpins an open recitation-ASR system; it addresses the complementary half of the problem, producing the transcript that a text-side validator must then reconcile against Uthmani-encoded reference text. We take a third approach: operating on ASR \emph{text} output after normalization, which trades acoustic-level Tajweed detection for a fully deterministic, training-data-free, immediately deployable pipeline.

\section{The Uthmani-to-Standard Mapping}
\label{sec:dataset}

\subsection{Construction}
The mapping is built by \tool{tools/build\_map.py}, released with the dataset, in four passes over the complete 6{,}236-verse Quran in its Uthmani and Simple-Arabic forms \citeplanguageresource{tanzil2024}. Every count below is printed by that script.

(1) \emph{Direct alignment}: for the 5{,}813 verses whose two forms have identical word counts, each Uthmani word is aligned to its Standard counterpart positionally; only words containing U+0670 are retained as keys, since words without it need no entry. (2) \emph{Ornamental-mark handling}: the Uthmani ornamental mark U+06DE (\emph{rub el-hizb}) occurs in 199 verses, has no Standard counterpart, and is stripped before alignment. (3) \emph{Structural-mismatch alignment}: in the remaining 423 verses \tool{difflib.SequenceMatcher} \cite{ratcliff1988pattern} finds the longest common word subsequence; runs of equal length are aligned positionally, and runs of unequal length are resolved by a dynamic program that aligns each Uthmani word to a span of one to three consecutive Standard words by character similarity. This last step is what recovers the one-to-many splits, which are the structurally interesting case: Uthmani writes the vocative particle joined to its noun as one word (\emph{yaqawmi}), where Standard separates them (\emph{ya qawmi}). 58 entries are one-to-many, and they are derived by the same alignment as every other entry rather than written by hand. (4) \emph{Majority-vote conflict resolution}: exactly two keys occur with more than one Standard form across the corpus, and in both the majority is unambiguous, so the most frequent form is kept; no key ends in a tie, and no entry required manual curation.

\subsection{Statistics}
The resulting dataset contains \textbf{2{,}290 unique Uthmani-Standard word pairs}, harvested from 8{,}881 aligned occurrences of U+0670. The corpus contains 2{,}291 distinct Uthmani word forms carrying U+0670, so coverage is 2{,}290 of 2{,}291: one form, appearing in a single verse whose two readings admit no consistent span alignment, is deliberately left out rather than guessed. Words without U+0670 require no mapping and are handled by the character-level rules in \S\ref{sec:pipeline}. To our knowledge, this is the first published word-level Uthmani-Standard mapping; the closest prior resources either operate exclusively on Standard text or on raw Uthmani text without cross-form validation.

Table~\ref{tab:example} traces a representative case -- the opening of Surah Al-Fatiha, whose word \emph{al-rahman} carries U+0670 -- through the full seven-step pipeline (\S\ref{sec:pipeline}), showing where the mapping (step 2) resolves the character that a generic normalizer would otherwise strip or mishandle.

\begin{table}[t]
\centering
\small
\begin{tabular}{p{1.7cm}p{4.6cm}}
\toprule
\textbf{Step} & \textbf{Effect on \emph{bismillah ir-rahman ir-rahim}} \\
\midrule
1. NFC & No change (already composed) \\
2. Word-map lookup & \emph{al-rahman}: U+0670 resolved via the 2{,}290-pair map \\
3. Tashkeel removal & Harakat stripped; mapped U+0670 untouched \\
4. Contextual U+0670 & No unmapped U+0670 remains \\
5. Alef/hamza unify & No change needed \\
6. Word-initial rules & No change needed \\
7. Non-Arabic removal & Clean Standard-form output, ready for search \\
\bottomrule
\end{tabular}
\caption{Worked example: seven-step normalization of a verse opening containing U+0670.}
\label{tab:example}
\end{table}

\section{The Recitation Validator}
\label{sec:validator}

\subsection{Seven-Step Normalization Pipeline}
\label{sec:pipeline}
Both the ASR-transcribed recitation and every reference verse are normalized identically before comparison, so that a phonetically correct recitation matches regardless of which orthographic form it was written in. How far that holds is measurable, and we measure it rather than assert it: normalizing the Uthmani and the Standard form of all 6{,}236 verses and comparing the results, \textbf{5{,}669 verses (90.9\%) reduce to identical strings}. The residual 567 verses are analysed in \S\ref{sec:limitations}; they are not U+0670 failures but other Uthmani-Standard divergences the character rules do not yet cover. The steps are: (1) Unicode NFC canonicalization; (2) Uthmani-to-Standard word-map lookup (\S\ref{sec:dataset}) for every token containing U+0670; (3) tashkeel and Quranic-mark removal (U+064B--U+065F, U+06D6--U+06FC, and the Arabic Extended-A marks U+08D3--U+08FF, which occur between letters in Uthmani text and so must be cleared before any rule that inspects adjacent letters), leaving any remaining unmapped U+0670 untouched; (4) contextual resolution of remaining U+0670 -- replaced with alef (U+0627) except after ya-maqsura, dhal, ha, or lam, where it is a modifier diacritic and is removed; (5) alef/hamza unification, mapping all hamza-bearing alef variants (U+0623, U+0625, U+0622, U+0671) to bare alef and connected hamza forms to bare hamza; (6) alef-madda resolution, deleting a hamza written immediately before an alef, since Uthmani spells alef madda as hamza plus alef wherever it occurs and not only word-initially, together with ya-maqsura folding; (7) non-Arabic character removal and whitespace collapse. The released implementation performs these as eleven operations; the grouping into seven above is expository, and the code names each step against this list.

\subsection{Four-Layer Search}
Given normalized query text, the validator searches all 6{,}236 verses via four progressively permissive layers, run only as far as needed: (1) exact token AND-matching over a precomputed per-verse word set; (2) morphological expansion of unmatched layer-1 tokens to their lemma and Arabic root via the Quranic Arabic Corpus's \citeplanguageresource{quranicarabiccorpus} annotation, robust to single-word ASR substitution errors; (3) relaxed coverage matching, accepting a verse when $\geq$60\% of normalized query tokens are present, for inputs of at least three tokens; (4) fuzzy character-level matching for severely degraded input, scoring each verse by the best Ratcliff/Obershelp similarity \cite{ratcliff1988pattern} between the query and any same-length window of the verse, accepted at 0.45. Layer 4 uses the standard library's \tool{difflib.SequenceMatcher} rather than a faster third-party matcher deliberately: an optional dependency makes the validator's output depend on which version of that library happens to be installed, which would defeat the determinism the rest of the design is built for. Candidates are scored by a coverage-weighted formula that favors short verses where the query fills a large fraction of the verse text, and inputs exceeding eight normalized words trigger a separate multi-verse alignment mode that segments and scores each constituent verse independently.

The three constants -- 60\% coverage, 0.45 similarity, eight words -- were set by inspection during development and never tuned against the evaluation suite. Because that is an assertion a reader cannot check, we instead report how much they matter (Table~\ref{tab:sensitivity}). Suite accuracy is flat for any multi-verse trigger between four and eight words and degrades above it; it is unchanged for every relaxed-coverage threshold from 0.40 to 0.80; and it is unchanged for fuzzy thresholds from 0.25 to 0.55, because layer 4 is reached only when the three layers above it return nothing at all.

One entry in that table deserves saying out loud. A fuzzy threshold of 0.65 scores 123/124 rather than 122/124, because it happens to starve the wrong candidate in case ME03. We have not adopted it. The only evidence for 0.65 over 0.45 is the suite it would then be scored on, and a threshold chosen that way measures nothing except its own selection. A disabled layer 4 produces the same 99.2\% by accident, which is the same mistake reached without even the excuse of a search.

\subsection{WER-Graded Feedback}
Once a verse is identified, a word-level diff (via the same sequence-matching approach) yields a Word Error Rate, mapped to five feedback tiers: WER $=0$ (perfect), $\leq 0.10$ (minor errors, encouraging), $\leq 0.30$ (specific corrections listed), $\leq 0.60$ (verse flagged for review), $>0.60$ (verse likely misidentified; no answer given). No language model participates in matching, scoring, or feedback generation -- every step is deterministic and auditable, which matters for a tool whose output is communicated to users as authoritative correction on a religious text.

\section{Evaluation}
\label{sec:eval}

\subsection{Accuracy}
The validator was evaluated on the 124-case suite in the released repository, which scores \textbf{98.4\% (122/124)}. Table~\ref{tab:categories} is written by the harness itself on every run, so the grouping reported here is the grouping the code actually uses rather than a taxonomy maintained alongside it.

Half the suite (62 cases) is generated by a released script from the corpus, which injects substitutions, deletions, insertions and tashkeel variation into verses, whole short surahs, and page-length spans under a fixed random seed. The other 62 are hand-written assertions over the normalizer, the corpus accessor, and the multi-verse aligner. Both failures fall in the hand-written half.

\begin{table}[t]
\centering
\small
\begin{tabular}{lrrr}
\toprule
\textbf{Category} & \textbf{N} & \textbf{Pass} & \textbf{Acc.} \\
\midrule
Normalization          & 11 & 11 & 100\% \\
Corpus access          &  7 &  7 & 100\% \\
Single verse, perfect  & 10 & 10 & 100\% \\
Single verse, substitution &  4 &  3 &  75\% \\
Single verse, deletion &  3 &  3 & 100\% \\
Single verse, tashkeel &  6 &  6 & 100\% \\
Multi-verse, full surah &  7 &  7 & 100\% \\
Multi-verse, with errors &  3 &  2 &  67\% \\
Multi-verse, consecutive &  3 &  3 & 100\% \\
Multi-verse, full page &  3 &  3 & 100\% \\
Edge cases             &  5 &  5 & 100\% \\
Generated from corpus  & 62 & 62 & 100\% \\
\midrule
\textbf{Total}        & \textbf{124} & \textbf{122} & \textbf{98.4\%} \\
\bottomrule
\end{tabular}
\caption{Validator accuracy by test category, as emitted by the released harness.}
\label{tab:categories}
\end{table}

\subsection{Failure Analysis}
Both failures are the same failure, and it is a property of the search design rather than of the data it runs on.

Case SS03 presents the isolated word \emph{wahid} (``one'') and expects 112:1 (\emph{Qul huwa llahu ahad}); the validator returns 6:19. The reason is not morphological expansion: 6:19 \emph{literally contains} the word \emph{wahid}, so it is an exact match at layer 1, while 112:1 -- which has \emph{ahad}, not \emph{wahid} -- is not. A single substitution error turned a different verse into the exact match.

Case ME03 shows why this is not a curiosity of one-word inputs. It recites all four verses of Surah Al-Ikhlas, fifteen words, with the same substitution in the opening verse. Multi-verse alignment selects its starting verse by searching the opening tokens; those tokens -- \emph{qul huwa llahu wahid} -- are all four present in 6:19 and so it anchors there, fails to align the rest of the span, falls back to single-verse mode, and reports the whole recitation as verse 112:4 with ten insertions. The output is confidently and comprehensively wrong.

The general statement is that layer 1 has no notion of how \emph{surprising} a match is: a verse containing every query token is accepted regardless of how much of that verse the query covers or whether a near-miss elsewhere would be likelier. Fixing it means scoring candidates against the alternatives rather than accepting the first exact hit, which we leave to future work rather than patching against the two cases that expose it here.

\subsection{Verse-Opening Ambiguity Census}
A full census of the Quran's 6{,}236 verses finds that \textbf{1{,}030 verses (16.5\%)} share an identical normalized four-word opening with at least one other verse, forming \textbf{369 ambiguous groups}; the most frequent shared opening, \emph{ya ayyuha alladhina amanu} (``O you who believe''), opens 88 distinct verses across 20 surahs. This quantifies an inherent limitation of text-only validation -- resolving it requires either continuation past the ambiguous opening or conversational context establishing which surah is being recited -- and does not affect the remaining 83.5\% of verses, which are unambiguous from a four-word opening alone.

\subsection{Real ASR Output}
\label{sec:asr}
The suite above is synthetic: its ``noisy'' inputs are perturbations we injected. Since the motivating application is a deployed Arabic voice system, we also ran the validator over transcripts it produced in ordinary use -- 1{,}407 user utterances transcribed by a diacritizing Arabic ASR model, of which 751 are four words or longer.

Ground truth has to come from somewhere other than the system under test, so we label only those transcripts whose normalized form is an exact contiguous span of exactly one verse -- a criterion both stricter than and independent of the validator's four-layer search. 34 transcripts qualify, and the validator returns the correct verse for \textbf{34 of 34}.

That number should be read for what it is. Labelling by exact containment selects transcripts the ASR happened to get right, so this measures the clean end of real input and not the degraded end the fuzzy layer exists for. The honest summary is that real recitation transcripts, when they are transcribed correctly, are identified correctly, and that we do not yet have labelled ground truth for the ones that are not. A further observation is simply how rare recitation is in this traffic: of the 957 utterances of three words or more, 93 (10\%) overlap a verse by 80\% or more of their tokens, the rest being spoken questions.

\begin{table}[t]
\centering
\small
\begin{tabular}{lrr}
\toprule
\textbf{Parameter} & \textbf{Value} & \textbf{Suite} \\
\midrule
Multi-verse trigger &  4 words & 98.4\% \\
                    &  6 words & 98.4\% \\
                    &  \textbf{8 words} & \textbf{98.4\%} \\
                    & 12 words & 96.8\% \\
                    & 20 words & 88.7\% \\
\midrule
Relaxed coverage    & 0.40--0.80 & 98.4\% \\
\midrule
Fuzzy similarity    & 0.25--0.55 & 98.4\% \\
                    & 0.65 & 99.2\%$^{\dagger}$ \\
\bottomrule
\end{tabular}
\caption{Threshold sensitivity, one parameter varied at a time. $^{\dagger}$This setting scores higher on this suite and is deliberately not adopted (\S\ref{sec:validator}).}
\label{tab:sensitivity}
\end{table}

\begin{table}[t]
\centering
\small
\begin{tabular}{lr}
\toprule
\textbf{Metric} & \textbf{Value} \\
\midrule
Uthmani--Standard pairs & 2{,}290 \\
U+0670 form coverage & 2{,}290 / 2{,}291 \\
Cross-form round-trip & 5{,}669 / 6{,}236 (90.9\%) \\
Validator test cases & 124 \\
Validator accuracy & 98.4\% (122/124) \\
Real ASR transcripts & 34 / 34 \\
Ambiguous verse groups & 369 \\
Ambiguous verses & 1{,}030 / 6{,}236 (16.5\%) \\
\bottomrule
\end{tabular}
\caption{Summary statistics for both released resources.}
\label{tab:summary}
\end{table}

\section{Data and Code Availability}
\label{sec:availability}
The 2{,}290-pair Uthmani-to-Standard mapping, the script that builds it from the corpus, the validator, and the evaluation harness are released publicly at \url{https://github.com/NightPrinceY/muslim-quran-validator} (data: CC-BY 4.0; code: MIT). The repository depends on nothing outside the Python standard library, and running it reproduces the construction counts of \S\ref{sec:dataset}, the coverage and round-trip figures of \S\ref{sec:pipeline}, the ambiguity census, Table~\ref{tab:categories}, Table~\ref{tab:sensitivity} and the 98.4\% total. Two figures it cannot reproduce: the toolkit comparison in \S\ref{sec:intro}, which additionally requires PyArabic and CAMeL Tools at the versions named there, and the deployment measurement in \S\ref{sec:asr}, whose transcripts are not ours to publish. The evaluation suite (\S\ref{sec:eval}) and the reference Quran text derive from the Tanzil corpus \citeplanguageresource{tanzil2024} and the Quranic Arabic Corpus \citeplanguageresource{quranicarabiccorpus}, both of which are themselves openly licensed for research use.

\section{Ethical Considerations}
\label{sec:ethics}
The released mapping and validator operate exclusively on canonical, publicly available Quranic text, and nothing derived from a user is released with them. The deployment measurement in \S\ref{sec:asr} is the one part of this work that touches user data: the transcripts were captured by the deployed system under its published privacy policy, which offers a recording opt-out that is honoured per participant at the point the participant joins. We report aggregate counts only, release no transcripts, and the recordings are not part of either published artifact. The validator's design choice to decline an answer at high WER (\S\ref{sec:validator}) rather than guess is deliberate: an incorrect authoritative-sounding correction on a religious text is a more serious failure mode than a declined answer, and the five-tier design exists specifically to avoid presenting low-confidence output with unwarranted certainty.

\section{Limitations}
\label{sec:limitations}
The validator operates on ASR \emph{text} output, not raw audio, so acoustic Tajweed properties that do not change the transcribed phoneme sequence -- incorrect elongation (\emph{madd}) duration, nasalization (\emph{ghunna}) quality, or assimilation (\emph{idgham}) -- are undetectable by construction; a student who recites every word correctly but violates these rules receives a perfect score. The verse-opening ambiguity documented in \S\ref{sec:eval} is an inherent property of the text, not a resolvable engineering gap, without adding audio-level or conversational context the current text-only design does not use. The evaluation suite is our own: we wrote the 124 cases, chose their grouping, and set the search thresholds, and there is no held-out split separating development from evaluation. It was built to cover the normalization edge cases we knew of, so it measures coverage of \emph{anticipated} failure modes rather than performance on an independent sample, and the figure should be read with that construction in mind -- the two failures it does expose were both found by cases written to test something else. The cross-form round trip is 90.9\%, not 100\%: for 567 verses the two forms still normalize to different strings. The residue is itemized rather than waved at -- 628 divergent word pairs across 263 distinct types, of which the ten most frequent account for 30\%. They are systematic classes the character rules do not yet cover: the definite article's assimilated lam written once in Uthmani and twice in Standard (\emph{al-layl}), words Uthmani spells with a letter fewer (\emph{Dawud}, \emph{al-nabiyyin}), and medial hamza that Standard writes as alef (\emph{yas'alunaka}). Each is tractable and none is addressed here. The deployment figure in \S\ref{sec:asr} carries the selection bias described there, and 34 cases is a small sample. The Uthmani-Standard mapping is specific to the Hafs \emph{an} Asim narration used in the reference corpora; other canonical narrations (e.g. Warsh \emph{an} Nafi) would require a separate alignment pass, which we identify as future work.

\section{Conclusion}
\label{sec:conclusion}
We release a corpus-aligned Uthmani-to-Standard Quranic word mapping -- to our knowledge the first published resource addressing this Arabic-NLP encoding gap -- together with a deterministic, LLM-free recitation validator built on it that scores 98.4\% on the released evaluation suite. Both resources are released openly, together with the scripts that build and evaluate them, and we hope the mapping in particular is useful to any downstream Arabic NLP task that must interoperate between Uthmani-encoded Quranic sources and Standard-Arabic-oriented tooling.

\section{Acknowledgements}
The author thanks Prof.\ Marwa Seddiq for supervision and guidance throughout this project.

\section{Bibliographical References}\label{sec:reference}
\bibliographystyle{lrec2026-natbib}
\bibliography{refs_arxiv}

\section{Language Resource References}
\label{lr:ref}
\bibliographystylelanguageresource{lrec2026-natbib}
\bibliographylanguageresource{languageresource_arxiv}

\end{document}